\documentclass[letterpaper]{article} 
\usepackage{aaai23}  
\usepackage{times}  
\usepackage{helvet}  
\usepackage{courier}  
\usepackage[hyphens]{url}  
\usepackage{graphicx} 
\usepackage{natbib}  
\usepackage{caption} 
\usepackage{algorithm}
\usepackage{subfigure}
\usepackage{multicol, multirow}
\usepackage{makecell}
\usepackage{amsthm}
\usepackage{amsmath}
\usepackage{amssymb}
\usepackage[noend]{algpseudocode}

\newtheorem{Def}{Definition}

\usepackage{newfloat}
\usepackage{listings}
\DeclareCaptionStyle{ruled}{labelfont=normalfont,labelsep=colon,strut=off} 
\floatstyle{ruled}
\newfloat{listing}{tb}{lst}{}
\floatname{listing}{Listing}

\usepackage{color}
\definecolor{codegreen}{rgb}{0,0.6,0}
\definecolor{codegray}{rgb}{0.5,0.5,0.5}
\definecolor{codepurple}{rgb}{0.58,0,0.82}
\definecolor{backcolour}{rgb}{0.95,0.95,0.92}
\definecolor{inlinecode}{rgb}{0.8,0.8,0.8}
 
\lstdefinestyle{pythonstyle}{
    backgroundcolor=\color{backcolour},   
    commentstyle=\color{codegreen},
    keywordstyle=\color{magenta},
    numberstyle=\tiny\color{codegray},
    stringstyle=\color{codepurple},
    basicstyle=\footnotesize,
    breakatwhitespace=false,         
    breaklines=true,                 
    captionpos=b,                    
    keepspaces=true,                 
    numbers=left,                    
    numbersep=5pt,                  
    showspaces=false,                
    showstringspaces=false,
    showtabs=false,                  
    tabsize=2
}
\title{NAS-LID: Efficient Neural Architecture Search with Local Intrinsic Dimension}
\author{
    Xin He,\textsuperscript{\rm 1,4}
    Jiangchao Yao, \textsuperscript{\rm 2,3}
    Yuxin Wang, \textsuperscript{\rm 1}
    Zhenheng Tang, \textsuperscript{\rm 1}\\
    Ka Chun Cheung, \textsuperscript{\rm 1,4}
    Simon See, \textsuperscript{\rm 2,4,6,7}
    Bo Han, \textsuperscript{\rm 1}
    Xiaowen Chu, \textsuperscript{\rm 1,5}\thanks{Corresponding author: xwchu@ust.hk}
}
\affiliations{
    \textsuperscript{\rm 1} Hong Kong Baptist University
    \textsuperscript{\rm 2} Shanghai Jiao Tong University \\
    \textsuperscript{\rm 3} Shanghai AI Laboratory
    \textsuperscript{\rm 4} NVIDIA AI Tech Center \\
    \textsuperscript{\rm 5} The Hong Kong University of Science and Technology (Guangzhou) \\
    \textsuperscript{\rm 6} Mahindra University
    \textsuperscript{\rm 7} Coventry University
}

\usepackage{bibentry}

\begin{document}

\maketitle

\begin{abstract}

One-shot neural architecture search (NAS) substantially improves the search efficiency by training one supernet to estimate the performance of every possible child architecture (i.e., subnet). However, the inconsistency of characteristics among subnets incurs serious interference in the optimization, resulting in poor performance ranking correlation of subnets. Subsequent explorations decompose supernet weights via a particular criterion, e.g., gradient matching, to reduce the interference; yet they suffer from huge computational cost and low space separability. In this work, we propose a lightweight and effective local intrinsic dimension (LID)-based method \textit{NAS-LID}. NAS-LID evaluates the geometrical properties of architectures by calculating the \emph{low-cost} LID features layer-by-layer, and the similarity characterized by LID enjoys \emph{better separability} compared with gradients, which thus effectively reduces the interference among subnets. Extensive experiments on NASBench-201 indicate that NAS-LID achieves superior performance with better efficiency. 
Specifically, compared to the gradient-driven method, NAS-LID can save up to 86\% of GPU memory overhead when searching on NASBench-201. We also demonstrate the effectiveness of NAS-LID on ProxylessNAS and OFA spaces. Source code: \url{https://github.com/marsggbo/NAS-LID}.

\end{abstract}

\section{Introduction}

Neural architecture search (NAS)~\cite{nas-survey,he2021automl} has been widely used to discover models automatically in various tasks~\cite{he2021covidnet,liu2019autodeeplab,ying2022eagan,he2022emars}. Vanilla NAS~\cite{nas2016,amoebanet} trains and evaluates each architecture separately, which obtains the true performance of all searched architectures at the cost of substantial computations. One-shot NAS~\cite{enas,liu2018darts} drastically reduces the cost by training only one supernet as an estimator of the performance of all subnets in the search space. However, the subnets with inconsistent characteristics interfere with each other during training in the shared supernet, resulting in inaccurate estimation~\cite{bender2018understanding}.

\begin{figure}
    \centering
    \includegraphics[width=0.39\textwidth]{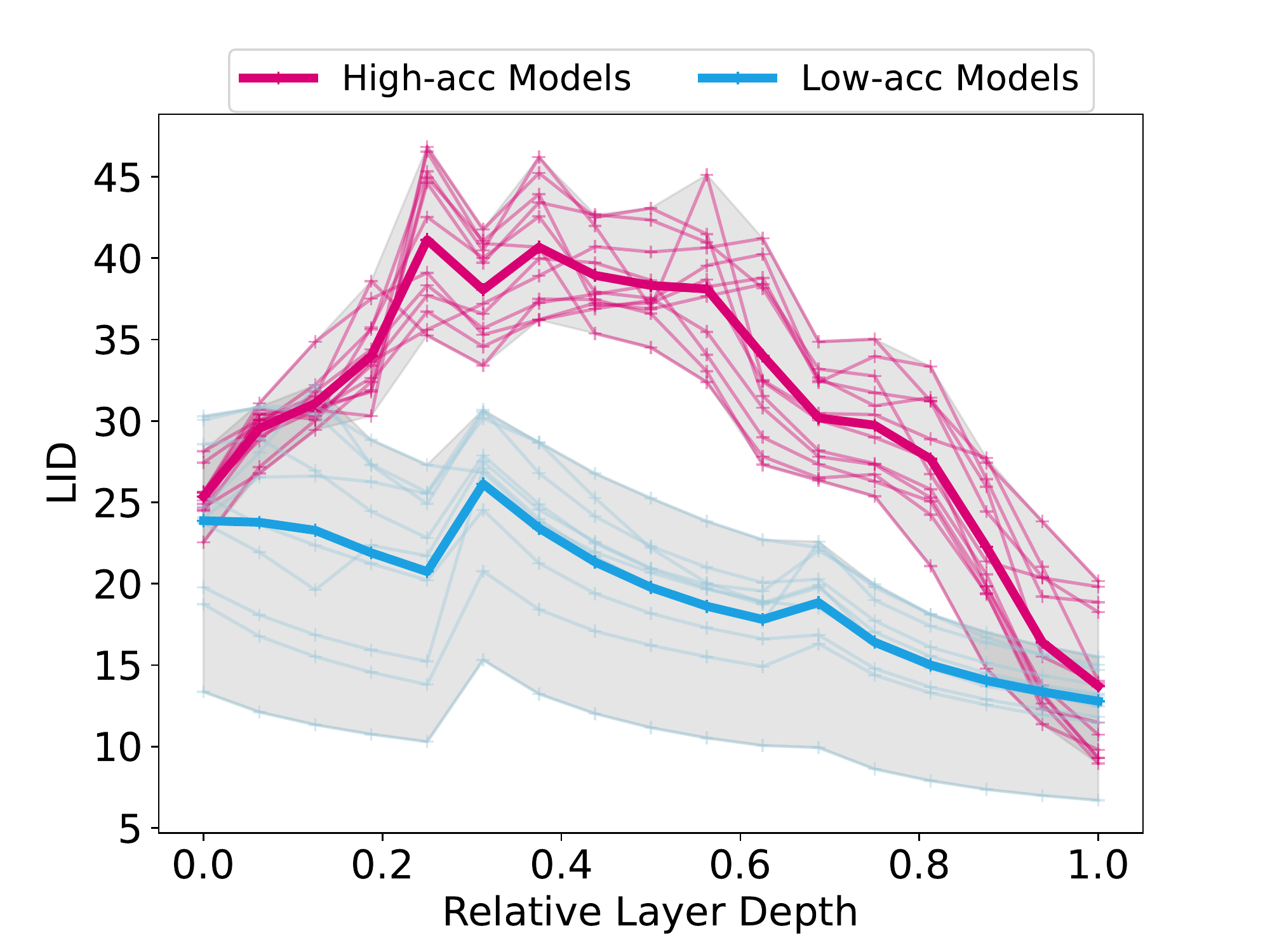}
    \caption{Layer-wise LID of sampled architectures in NASBench-201. Each curve indicates a model. The LID profiles of high-accuracy ($>$92\%) models are arch-shaped (i.e., first increasing and then decreasing), while those of low-accuracy ($<$86\%) models show an approximately monotonically decreasing trend.}
    \label{fig:LID-char}
\end{figure}

Recent works to reduce the interference among subnets can be categorized into two groups. One is to treat all subnets as student networks and use a well-designed teacher network to force the similarity of their layer-wise outputs~\cite{pinas,MAGIC,bashivan2019teachernas}. However, \citet{searchtodistill} empirically showed that the optimal student would be different under different teachers even trained on the same task and dataset. In other words, choosing a different teacher will lead to quite different results for the performance ranking of subnets. Instead, few-shot NAS~\cite{fewshotNAS,gmnas} aggregates the subnets with aligned inputs and outputs into the same sub-supernet by splitting the supernet into multiple sub-supernets. For example, GM-NAS~\cite{gmnas} splits the supernet via the gradient similarity between sub-supernets, which achieves state-of-the-art (SOTA) performance. However, the gradients are high-dimensional and sparse data that inevitably encounter the curse of dimensionality on space separability. Our experiments in Sec. \ref{sec:nb201-exp} show that the gradient similarities among sub-supernets are very close, leading to low separability. Besides, GM-NAS requires the forward and backward processes for each sub-supernet before similarity calculation, incurring huge GPU memory occupation and high computational complexity.

To address the issue, we explore a different criterion, namely local intrinsic dimension (LID)~\cite{houleLID1}, which measures the minimal number of parameters to describe the data representation learned from the model. We randomly sample multiple models from NASBench-201~\cite{nb201} and train them from scratch to obtain accurate LID estimations of all layers. Fig.   \ref{fig:LID-char} presents LIDs against relative depth, where the LID is orders of magnitude smaller than the number of parameters per layer. Notably, the models with higher accuracy have arch-shaped LID profiles, i.e., LID first increases and then decreases, which has been similarly observed in ~\cite{id-profile}. On the other hand, we show that the LID profiles of the models with lower accuracy are monotonically decreasing, except for a slight increase at relative depths of 0.3 and 0.7. If the LIDs of two data are close, it means that they are close in the low-dimensional manifold, as well as in high-dimensional space. Models with similar LID profiles tend to have similar layer-wise outputs. The observation implies that LID profiles can capture the geometrical properties of architectures and determine the intrinsic training dynamics.



Thus, we propose a new method \textit{NAS-LID}, which leverages LID to characterize the similarity among architectures for NAS. Each time we compare all unpartitioned layers to select the one with the highest partition score based on LID (see Sec. \ref{sec:ss-partition}). Then, we partition the candidate operations on the selected layer into two sub-supernets. Iterating this partition allows us to get smaller sub-supernets. Thanks to the advantages of LID, we both effectively avoid the curse of dimensionality and reduce the computational cost in this procedure. In a nutshell, the contribution of this work can be summarized as follows.




\begin{enumerate}
    \item We are the first to provide a proof-of-concept for the potential application of LID in splitting one-shot supernet, and discover that the LID-based characterization exhibits better space separability and higher performance ranking scores than the gradient counterpart. 
    \item Compared to the gradient-based split scheme, we propose a novel method NAS-LID, which can significantly reduce the GPU memory overhead (saving up to 86\% of GPU memory overhead on NASBench-201) and guarantee the superior performance.
    \item We demonstrate the effectiveness of NAS-LID by conducting extensive experiments on multiple search spaces (NASBench-201, ProxylessNAS, and OFA) and datasets (CIFAR10, CIFAR100, and ImageNet).
\end{enumerate}

\section{Related Work}

\subsection{Neural Architecture Search}
Vanilla NAS~\cite{nas2016,nasnet,amoebanet} finds the best architecture by training all sampled architectures to compare their performance, which requires vast computational resources. One-shot NAS~\cite{enas,liu2018darts,cai2018proxylessnas} represents the search space as a super network (i.e., supernet), where each possible architecture is regarded as a subnet. In this way, we can train only one supernet to estimate the performance of all subnets. Since the weights of all subnets are coupled, they interfere with each other during the training process~\cite{bender2018understanding}, resulting in inaccurate performance predictions of subnets by the supernet.

There are two main types of methods to reduce interference. One aligns the outputs of all layers of the subnets with those of the teacher network~\cite{bashivan2019teachernas,pinas,MAGIC} via the knowledge distillation approach~\cite{kd_hinton}. For example, \citet{MAGIC} selected a top-performing subnet from the search space as the teacher network to align other subnets. Nevertheless, \citet{searchtodistill} empirically demonstrated that different student networks are not equally capable of distilling knowledge from different teacher networks. Teacher networks are likely to introduce innate biases, leading to unfair assessments of model performance. The other promising solution is to split the supernet into multiple sub-supernets~\cite{fewshotNAS,gmnas,kshotNAS}, where the architectures with similar characterization are in the same sub-supernet. \citet{fewshotNAS} exhaustively divided the supernet into hundreds of sub-supernets,leading to high resource consumption. To reduce the cost, GM-NAS~\cite{gmnas} used the gradient information at the shared part of sub-supernets to partition the supernet. However, GM-NAS suffers from the curse of dimensionality, since the gradients are high-dimensional and sparse data.

\subsection{Intrinsic Dimension}


Intrinsic dimension (ID), the least number of parameters to describe data at minimum information loss, reflects the necessary degrees of freedom of a high-dimensional data space. To efficiently analyze complex and sparse high-dimensional data in deep neural networks, researchers have designed different ID estimators~\cite{id-benchmark,houleLID1} for different tasks. For example, \citet{xu2022fedcorr} use ID for adversarial detection to tackle heterogeneous label noise. \citet{li_id_2018_ICLR} leverage ID to estimate objective landscapes and show a method for approximating the model's training in a random subspace. Recently, with the development of large language models, \citet{id-explain-lmf} use ID to reduce the parameter sizes for fine-tuning. In this work, we use LID~\cite{houleLID1} to estimate ID because LID requires only an ordered list of neighbor distances to compute estimates, avoiding expensive matrix operations. We are the first to bring LID for NAS to build a promising criterion that helps achieve efficient NAS.

\begin{figure*}
    \centering
    \includegraphics[width=0.98\textwidth]{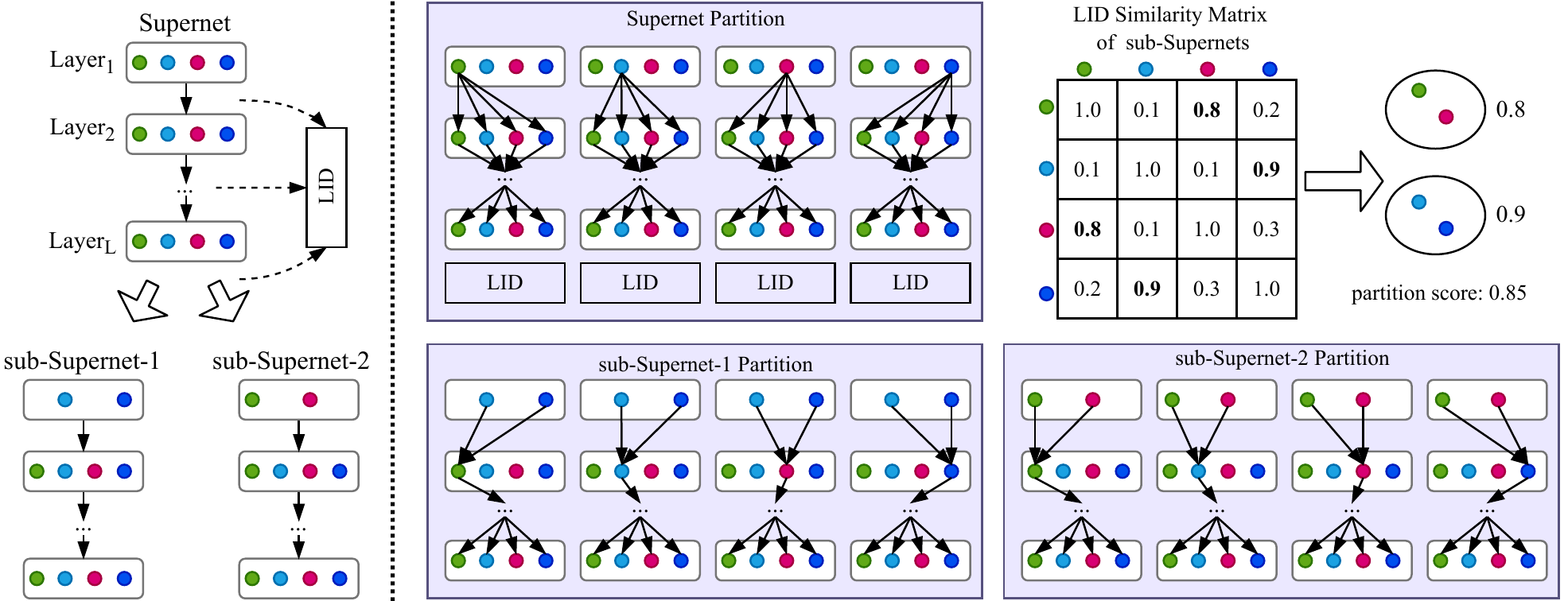}
    \caption{An illustration of the supernet partition method via local intrinsic dimension (LID). We compare all unpartitioned layers and only select the layer with the highest partition score to split the supernet. Taking the first layer as an example, we can split the operations into four different sub-supernets, each characterized by layer-wise LID. We calculate the similarity between LIDs of these sub-supernets and divide the operations into two groups by maximizing the sum of intra-similarity of two groups (i.e., partition score). Finally, we merge the operations of the same group to form two sub-supernets. We can obtain more sub-supernets by iterating the above steps on generated sub-supernets.}
    \label{fig:IDNAS}
\end{figure*}

\section{Methods}


In this work, we propose a lightweight and effective LID-based method, \textit{NAS-LID}, to show that LID can facilitate the separability of search space and thus improve the performance ranking correlation of subnets. In this section, we first introduce the LID and its estimation method. Then, we describe our proposed NAS-LID in two parts: 1) why and how we use LID for architecture characterization, and 2) how we split the supernet by LID similarity.

\subsection{Preliminary of Local Intrinsic Dimension}



LID~\cite{houleLID1} measures the ID of data representations in deep networks, without accessing the underlying global data distribution, which can be defined as follows.

\begin{Def}[Local Intrinsic Dimension~\cite{houleLID1}]
Given a random variable $\mathbf{R}$, denoting the distance from a reference sample to other samples. For any distance threshold $r$ such that the cumulative distance function $F_{\mathbf{R}}(r)$ is positive and continuously differentiable at $r$, then the LID of the reference sample at distance $r$ is given by

\begin{align}
\label{eq:lid1}
\begin{split}
\operatorname{LID}_{\mathbf{R}}(r) &\triangleq \lim _{\epsilon \rightarrow 0^+} \frac{\ln (F_{\mathbf{R}}((1+\epsilon) \cdot r) / F_{\mathbf{R}}(r))}{\ln ((1+\epsilon)\cdot r / r)} \\
&= \frac{r \cdot F_{\mathbf{R}}'(r)}{F_{\mathbf{R}}(r)}
\end{split}
\end{align}



\end{Def}


\noindent The last line of Eq.~\eqref{eq:lid1} gives a low-dimensional computation of the sub-manifold in the limit. A popular empirical estimation from~\cite{estimateLID} is maximum likelihood estimator (MLE). Specifically, given a reference data sample $x\sim \mathbf{X}$, where $\mathbf{X}$ represents the data distribution, the MLE estimator of the LID at $x$ is defined as follows:




\begin{equation}
\widehat{\mathrm{LID}}(x)=-\left(\frac{1}{k} \sum_{i=1}^{k} \ln \frac{r_{i}(x)}{r_{k }(x)}\right)^{-1}
\label{eq:lid-estimate}
\end{equation}

\noindent where $r_i(x)$ indicates the distance between $x$ and its $i$-th nearest neighbor within samples drawn from $\mathbf{X}$, i.e., $r_{k}(x)$ is the maximum distance from $x$ among its $k$ nearest neighbors. In our work, we use Euclidean distance and set $k=20$.

\subsection{The LID-based Characterization}

\subsubsection{Why do we use the LID-based Characterization?} 

In terms of space separability, model gradients used in previous SOTA GM-NAS are in a high-dimensional and sparse space. This is difficult to distinguish the difference between different gradients based on the Euclidean or cosine distance measure, making the partition inaccurate. In contrast, LID measures the dimensionality of the lower dimensional sub-manifold in which the high-dimensional data resides. As Fig.  ~\ref{fig:LID-char} shows, the high-dimensional outputs from each layer of the models in NASBench-201 fall in sub-manifolds with dimensions less than 50. Our experiments in Sec.~\ref{sec:ss-partition} show that our characterization can achieve much higher space separability than the gradient counterpart.

In terms of information bottleneck~\cite{tishby2000information,shwartz2017opening}, layer-wise LID profiles capture models' geometrical properties~\cite{lei2020geometric} and describe how information changes layer by layer. For arch-shaped LID profiles (red lines in Fig. \ref{fig:LID-char}), the initial increase of LID indicates that models are learning data by continuously mapping the input data to higher-dimensional manifolds, and the decreasing part is to prune features irrelevant for prediction. On the other hand, the monotonically decreasing LID profiles (blue lines in Fig. \ref{fig:LID-char}) indicate that these models fail to extract informative features at the initial layers, resulting in worse performance.

\subsubsection{Architecture LID Characterization}


Consider an architecture with $L$ layers, and let $X=[x_1,x_2,...,x_L]$ be the set of the feature representations of all layers, where $x_i\in \mathbb{R}^{b\times m_i}$, $b$ is the number of data (i.e., batch size), and $m_i$ is the size of the output representations in $i$-th layer. The LID of $i$-th layer representations (i.e., $x_i$) is calculated as follows

\begin{align}
    \mathrm{LID}(x_i)=\frac{1}{b}\sum_{j=0}^{b-1}\widehat{\mathrm{LID}}(x_{i}[j,:])
    \label{eq:LID-features}
\end{align}

\noindent where $x_i[j,:]$ is the $j$-th data sample of $x_i$. The architecture LID characterization is computed by stacking LIDs of all layers, namely $[\text{LID}_1, ..., \text{LID}_L]^T$, where $\text{LID}_i=\text{LID}(x_i)$.




\subsubsection{Sub-supernet LID Characterization}

Although we can directly separate the supernet by merging architectures (i.e., subnets) with high LID similarity, a standard NAS search space usually contains millions or even more subnets; thus, it is too expensive to compute LIDs of all subnets. For example, suppose a supernet $\mathcal{A}$ has $L$ searchable layers, each with $n$ candidate operations; it contains 1,048,576 subnets even when $n=4,L=10$. To improve efficiency, we split the supernet into a collection of sub-supernets. As Fig.   \ref{fig:IDNAS} shows, we can split the supernet by dividing the operations of one layer into different sub-supernets. In this way, the number of sub-supernets can be much less than the number of architectures, reducing the computational overhead. The LID characterization of the sub-supernet is similar to that of a single architecture. Without loss of generality, let's consider the LID of the $i$-th layer and denote by $\{x_{i}^{O_1},x_{i}^{O_2},...x_{i}^{O_n}\}$ the output features of $n$ candidate operations in $i$-th layer, by $I_i=\{I_{i}^{O_1},I_{i}^{O_2},...,I_{i}^{O_n}\}, I_{i}^{O_j}\in\{0,1\}$ the binary vector of candidate operations. Similar to Inception~\cite{inceptionv1}, the output of $i$-th layer in the sub-supernet is the sum of all selected operations' outputs, i.e., $z_i=\sum_{j=1}^n x_i^{O_j}\cdot I_{i}^{O_j}$; thus, the LIDs $\mathcal{L}$ of the sub-supernet $\mathcal{A}$ is defined as

\begin{align}
    \mathcal{L}_{\mathcal{A}}=[\operatorname{LID}(z_1),\operatorname{LID}(z_2),...,\operatorname{LID}(z_L)]^T
    \label{eq:lid-supernet}
\end{align}

\begin{figure}
    \centering
    \includegraphics[width=0.47\textwidth]{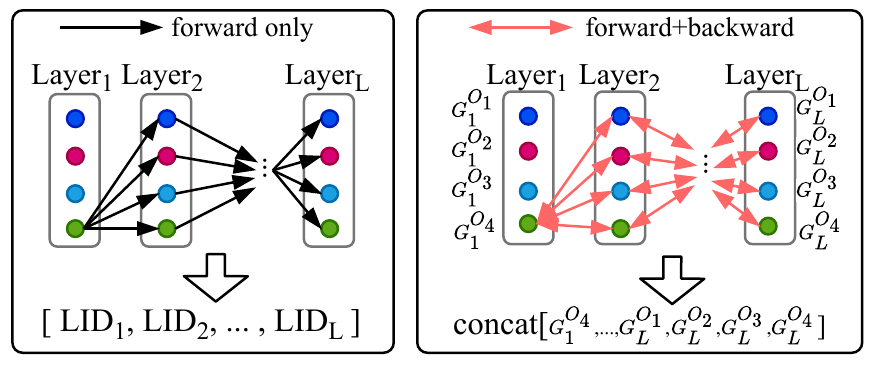}
    \caption{Overview of the $L$-layer sub-supernet characterization in our NAS-LID (left) and GM-NAS (right).  NAS-LID is more efficient as it requires only the forward process and $L$ parameters for characterization, while GM-NAS involves forward and backward processes and characterizes the sub-supernet by gradients of all selected operations.
    }
    \label{fig:IDvsGrad}
\end{figure}

\subsubsection{Complexity: Gradient vs. LID}


Fig. \ref{fig:IDvsGrad} presents difference in sub-supernet characterization between LID-based and gradient-based methods. Suppose we split a $L$-layer supernet into $n$ sub-supernets. NAS-LID only requires $n\times L$ parameters to characterize these sub-supernets, while GM-NAS requires approximately $n\times M$ parameters, where $M$ is the size of the supernet and $M\gg L$. 
In terms of GPU memory cost, NAS-LID involves only the forward process to compute layer-wise LID, while GM-NAS requires both forward and backward processes, and the overhead of backward is empirically twice that of the forward. Table \ref{tab:gpucost} compares the GPU memory overhead consumed by GM-NAS and NAS-LID on a single V100 GPU (32GB). We can see that NAS-LID is more efficient as it can save up to 86\% GPU memory on NASBench-201 with input size of $32\times3\times224\times224$. Thus, NAS-LID enables larger batch size for special needs of different computing tasks and scenarios \cite{yao2022edgesurvey}.


\begin{table}[!t]
    \centering
    \scalebox{0.95}{
        \begin{tabular}{c|ccc}\hline
          Input size &  Method & NASBench-201  & ProxylessNAS \\\hline\hline
         \multirow{2}{*}{\makecell{$128\times3$\\$\times32\times32$}}  & GM-NAS  &3,303  &  2,247  \\
           & NAS-LID &   \textbf{1,539 ($\downarrow53\%$)}  & \textbf{1,571 ($\downarrow 30\%$)}  \\\hline
         \multirow{2}{*}{\makecell{$32\times3$\\$\times24\times224$}}  & GM-NAS  & 24,073  & 6,707  \\
           & NAS-LID & \textbf{3,413 ($\downarrow 86\%$)}  & \textbf{2,015 ($\downarrow 70\%$)}  \\\hline
        \end{tabular}
    }
    \caption{Comparison of the GPU memory (MB) costs between GM-NAS and our proposed NAS-LID across different search spaces with different input sizes.}
    \label{tab:gpucost}
\end{table}

\subsection{LID-based Supernet Partition}\label{sec:ss-partition}

In this sub-section, we introduce how to split the supernet via LID. For simplicity, we use an example of a one-round supernet partition in Fig.   \ref{fig:IDNAS} to describe this procedure. Actually, NAS-LID consists of $T\geq 1$ rounds of partition, where the first round splits only the supernet, and the subsequent rounds split the sub-supernets generated in the previous rounds. After $T$ rounds of partition, we will get $2^T$ sub-supernets, each covering a different and non-overlapping region of the original search space. Alg. \ref{alg:lid-partition} details our LID-based supernet partition scheme. Without loss of generality, we introduce the partition on a sub-supernet in the $t$-th round. We denote by $\mathcal{A}=\{\mathcal{O}_1,\mathcal{O}_2,...,\mathcal{O}_L\}$ a sub-supernet, by $\mathcal{L}_{\mathcal{A}}$ the LID characterizations, where $\mathcal{O}_l=\{O_1,...,O_n\}$ is the set of $n$ candidate operations of the $l$-th layer and associated with a binary vector $I_l$ recording which operations have been deactivated (encoded by 0).




\subsubsection{Sub-supernet Similarity} 

As Fig.   \ref{fig:IDNAS} shows, the supernet can be divided into $n$ sub-supernets $\{\mathcal{A}_{O_1},...,\mathcal{A}_{O_n}\}$ by splitting the operations $(\mathcal{O}_{l})$ on the $l$-th layer; thus, the supernet partition is equivalent to operations partition. We use Eq. \ref{eq:lid-supernet} to get LIDs of these sub-supernets, i.e., $\{\mathcal{L}_{\mathcal{A}_{O_1}},...,\mathcal{L}_{\mathcal{A}_{O_n}}\}$, based on which we can obtain the $n\times n$ symmetric similarity matrix. via Eq. \ref{eq:lid-simlarity}. The similarity between two sub-supernets is defined as the reciprocal of the Euclidean distance of their LIDs.

\begin{align}
    S(\mathcal{A}_{O_i},\mathcal{A}_{O_j}) &= \frac{1}{\|\mathcal{L}_{\mathcal{A}_{O_i}},\mathcal{L}_{\mathcal{A}_{O_j}}\|_2+\epsilon}
    \label{eq:lid-simlarity}
\end{align}

\noindent where $\epsilon=10^{-6}$ avoids the denominator being 0. We have also explored Pearson distance, but our results in Sec. \ref{sec:sim-measure} show that Pearson distance is not suited for LID because the scale of LIDs is vital for characterization.

\subsubsection{Sub-Supernet Merging} Based on the similarity matrix, we can merge sub-supernets with high similarity by merging operations via the graph min-cut algorithm \cite{mincut}, which aims to maximize the partition score $\gamma$, i.e., the sum of intra-similarity of each group, as below:


\begin{align}
    \gamma=\operatorname*{max}_{\Gamma \subseteq   \mathcal{O}}& [\sum_{O,O'\in \Gamma} S(\mathcal{A}_{O},\mathcal{A}_{O'}) + \sum_{O,O'\in {\mathcal{O}\backslash\Gamma}} S(\mathcal{A}_{O},\mathcal{A}_{O'}) ]\notag\\
    \text{s.t. }& \lfloor n/2 \rfloor \leq|\Gamma|\leq \lceil n/2 \rceil 
    \label{eq:partition-score}
\end{align}

\noindent As a result, the candidate operations are divided into two groups $\{\Gamma,\mathcal{O}\backslash\Gamma\}$. In other words, the supernet $\mathcal{A}$ is split into two sub-supernets, i.e., $\mathcal{A}=\{\mathcal{A}_{\Gamma},\mathcal{A}_{\mathcal{O}\backslash\Gamma}\}$. After merging, we will update the binary vector $I$ of the partitioned layer for each generated sub-supernet. Next time, we will only split the unpartitioned layers (i.e., $I$ is an all-one vector) with the highest partition score. In this way, we can prevent the sub-supernets from being too fragmented and limit the number of sub-supernets, thus reducing the computational overhead.



\begin{algorithm}[!t]
    \caption{NAS-LID: LID-based Supernet Partition}
    \label{alg:lid-partition}
    \begin{algorithmic}[1] 
        \Require{$\mathcal{A}^0$: supernet, $T$: rounds of partition}
        \Ensure{$\Omega$: the set of partitioned sub-supernets}
        \Procedure{Main}{$\mathcal{A}^0,T$}
            \State $\Omega=\{\mathcal{A}^0\}$
            \For{$t\in[1,T]$}
                \State $\Omega^t=\{\}$
                \For{$\mathcal{A} \in \Omega$}
                    \State Warmup: training $\mathcal{A}$ as a one-shot supernet
                    \State $\Omega^t$.insert(\textit{SPLIT-SUPERNET}($\mathcal{A}$))
                \EndFor
                \State $\Omega=\Omega^t$
            \EndFor
            \State $\Omega\gets$ Finetuning all sub-supernets in $\Omega$
            \State \textbf{return} $\Omega$
        \EndProcedure
        \Procedure{Split-Supernet}{$\mathcal{A}$}
        
            \State $\mathcal{I}=\{I_1,...,I_L\}$ \Comment{binary vector of each layer in $\mathcal{A}$}
            \For{unsplitted $l$-th layer} \Comment{$sum(I_l)==|I_l|$}
                \State $\mathcal{O}_l=\{O_1,...,O_n\}$ \Comment{candidate operations}
                \State $\{\mathcal{A}_{O_1},...,\mathcal{A}_{O_n}\} \gets$ split $\mathcal{A}$ on the $l$-th layer 
                \State $\{\mathcal{L}_{\mathcal{A}_{O_1}},...,\mathcal{L}_{\mathcal{A}_{O_n}}\} \gets$ get LIDs via Eq. \ref{eq:lid-supernet}
                \State $\gamma \gets$ partition score via Eq. \ref{eq:lid-simlarity}$\sim$\ref{eq:partition-score}
            \EndFor
            \State split $\mathcal{A}$ on the layer with the best $\gamma^*$
            \State $\Gamma^*\gets$ best partitioned operation group
            \State \textbf{return} $\{\mathcal{A}_{\Gamma^*},\mathcal{A}_{\mathcal{O}\backslash\Gamma^*}\}$
        \EndProcedure
    \end{algorithmic}
\end{algorithm}

\section{Experiments}

In this section, we empirically verify the effectiveness of our proposed NAS-LID. We first conduct extensive experiments on NASBench-201 to compare the LID-driven and gradients-driven split schemes. We then evaluate the performance of NAS-LID on other open domain search spaces.

\subsection{NASBench-201}\label{sec:nb201-exp}

NASBench-201 is a public tabular architecture dataset, which builds a DARTS-like~\cite{liu2018darts} search space and provides the performance of 15,625 neural architectures on the CIFAR-10 and CIFAR-100 datasets~\cite{cifar10}. Each architecture is stacked with multiple cells, each sharing the same structure. As shown in Fig.   \ref{fig:nb201-sim-std}, a cell is represented as a directed acyclic graph (DAG) containing four nodes, and each edge has 5 predefined operations, i.e., None, Skip-connection, Conv-1$\times$1, Conv-3$\times$3, and Avgpool-3$\times$3. Thus, we can obtain five sub-supernets by splitting the operations on one edge. For example, if we split the edge (0-3), the five operations on this edge will be divided into five different sub-supernets that still retain five candidate operations on the remaining edges. We compare with GM-NAS~\cite{gmnas} in three aspects: separability, ranking correlation, and performance of derived architectures.


\begin{table*}[!ht]
    \centering
    \begin{tabular}{l|cc|cc|cc|cc} \hline
    \multirow{2}{*}{Method} & \multicolumn{2}{c|}{Top50} &\multicolumn{2}{c|}{Top100} & \multicolumn{2}{c}{Top150} \\ \cline{2-7}
    & Kendall & Spearman & Kendall & Spearman & Kendall & Spearman \\\hline
    \hline
    SPOS~\cite{spos} & 0.14$\pm$0.02 & 0.16$\pm$0.01 & 0.19$\pm$0.03 & 0.27$\pm$0.04 & 0.12$\pm$0.03 & 0.17$\pm$0.05 \\\hline
    GM-NAS~\cite{gmnas} & 0.34$\pm$0.09 & 0.48$\pm$0.11 & 0.21$\pm$0.04 & 0.29$\pm$0.07 & 0.23$\pm$0.03 & 0.31$\pm$0.06 \\\hline
    \textbf{NAS-LID (Ours)} & \textbf{0.47}$\pm$0.01 & \textbf{0.64}$\pm$0.03 & \textbf{0.35}$\pm$0.03 & \textbf{0.49}$\pm$0.04 & \textbf{0.34}$\pm$0.02 & \textbf{0.48}$\pm$0.03 \\\hline
    \end{tabular}
    \caption{Ranking correlation (the higher, the better) among top50/100/150 architectures of NASBench-201. Our LID-driven splitting scheme achieves a much higher ranking correlation with lower variance. }
    \label{tab:rank_perform}
\end{table*}

\subsubsection{Separability}\label{sec:separability} GM-NAS and our NAS-LID respectively split the supernet based on gradient and LID similarity. Here, we study how different these two criteria are to distinguish the sub-supernets, i.e., \textit{separability}. For the quantitative comparison, we first train the supernet for 50 epochs via Random Sampling with Parameter Sharing (RSPS)~\cite{rsps} and use the pretrained weights to calculate the separability score for each edge, formalized as follows.

\begin{align}
    D &= \sqrt{\frac{1}{2n}\sum_{i=1}^{n}\sum_{j=1}^{n}(s_{i,j}-\bar{s})^2} \notag\\
    \text{s.t.,} \,\, \bar{s}&=\frac{1}{2n}\sum_{i=1}^{n}\sum_{j=1}^{n}s_{i,j} \,\, \text{and} \,\,i\neq j 
\label{eq:dist}
\end{align}

\begin{figure}
    \centering
    \includegraphics[width=0.47\textwidth]{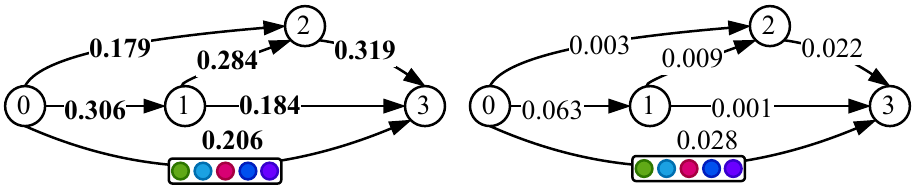}
    \caption{Comparison of separability score (the higher, the better) of each edge in NASBench-201 space between our NAS-LID (left) and GM-NAS (right).}
    \label{fig:nb201-sim-std}
\end{figure}

\noindent where $n$ is the number of candidate operations (i.e., 5 in NASBench-201), $s_{i,j}$ denotes the LID or gradient similarity between the two sub-supernets that only include operation $i$ and $j$, respectively. $\bar{s}$ is the average similarity of all pairs of sub-supernets. Intuitively, the lower the separability score, the higher the inter-similarity between different sub-supernets, and the harder it is to split them. Fig.   \ref{fig:nb201-sim-std} compares the separability score of each edge of NAS-LID and GM-NAS. We can see that the separability scores of all edges obtained by NAS-LID are two orders of magnitude higher than those obtained by GM-NAS, showing NAS-LID is more confident and certain in the partition.

\begin{figure}[!t]
    \centering
    \includegraphics[width=0.38\textwidth]{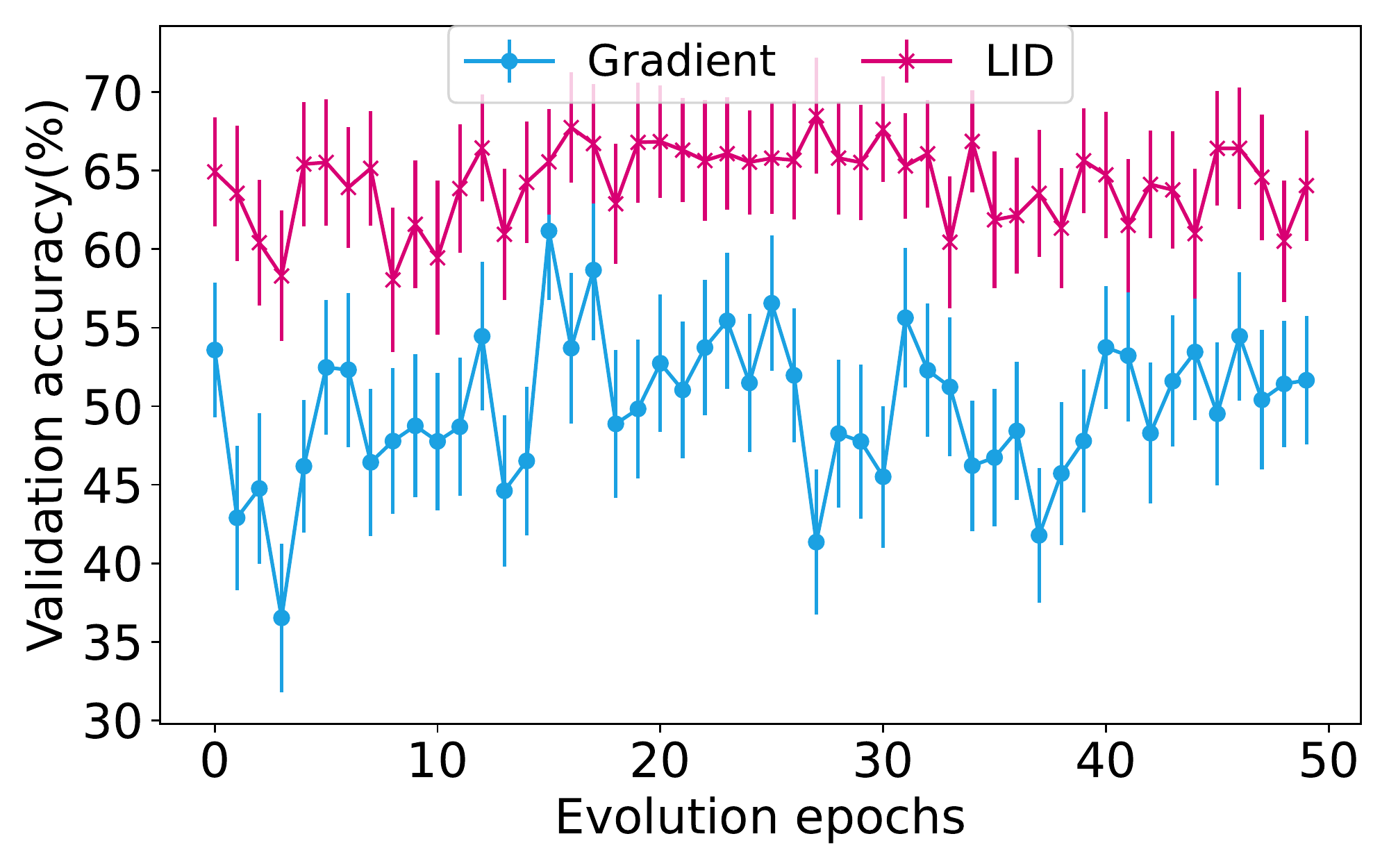}
    \caption{Validation accuracy of GM-NAS and our NAS-LID on NASBench-201 during the evolutionary search.}
    \label{fig:EASearchNB201}
\end{figure}

\begin{table*}[!ht]
\centering
\begin{tabular}{l|c|cc|cc}
\hline
\multirow{2}{*}{Method} & \multirow{2}{*}{Type} & \multicolumn{2}{c|}{CIFAR-10} & \multicolumn{2}{c}{CIFAR-100} \\ \cline{3-6} 
 & & validation & test & validation & test \\ \hline
 \hline
ResNet~\cite{resnet} &Manual & $90.83$ & $93.97$ & $70.42$ & $70.86$ \\ \hline
NASNet~\cite{nasnet} & Vanilla & $91.09 \pm 0.37$ & $93.85 \pm 0.37$ & $70.05 \pm 1.67$ & $70.17 \pm 1.61$ \\ \hline
ENAS~\cite{enas} & \multirow{7}{*}{One-shot} & $37.51 \pm 3.19$ & $53.89 \pm 0.58$ & $13.37 \pm 2.35$ & $13.96 \pm 2.33$ \\
DARTS~\cite{liu2018darts} &   & $39.77 \pm 0.00$ & $54.30 \pm 0.00$ & $15.03 \pm 0.00$ & $15.61 \pm 0.00$ \\
GDAS~\cite{gdas} &   & $90.00 \pm 0.21$ & $93.51 \pm 0.13$ & $71.14 \pm 0.27$ & $70.61 \pm 0.26$ \\
DSNAS~\cite{hu2020dsnas} &   & $89.66 \pm 0.29$ & $93.08 \pm 0.13$ & $30.87 \pm 16.40$ & $31.01 \pm 16.38$ \\
SETN~\cite{setn} & & $82.25\pm5.17$ & $86.19\pm4.63$ & $58.86\pm7.59$ & $56.87\pm7.77$ \\
PC-DARTS~\cite{xu2019pcdarts} &   & $89.96 \pm 0.15$ & $93.41 \pm 0.30$ & $67.12 \pm 0.39$ & $67.48 \pm 0.89$ \\
RSPS~\cite{rsps} &   & $84.16 \pm 1.69$ & $87.66 \pm 1.69$ & $59.00  \pm 4.60$ & $58.33 \pm 4.34$ \\\hline
FSNAS+RSPS~\cite{fewshotNAS} &  \multirow{3}{*}{Few-shot} & $85.40 \pm 1.28$ & $89.11 \pm 1.37$ & $58.59 \pm 3.45$ & $58.69 \pm 3.75$ \\
GM-NAS+RSPS~\cite{rsps} &   & $8 9 . 0 9 \pm 0 . 4 0$ & $9 2 . 7 0 \pm  0 . 5 3 $ & $6 8 . 3 6 \pm0 . 9 1 $ & $6 8 . 8 1 \pm 1 . 2 8$ \\
\textbf{NAS-LID+RSPS (ours)}  &  & \textbf{89.74 $\pm$ 0.37} & \textbf{92.90 $\pm$ 0.27} & \textbf{69.38 $\pm$0.36} & \textbf{69.39 $\pm$0.77 }  \\\hline
optimal & &   91.61 & 94.37 & 73.49 & 73.51 \\ \hline
\end{tabular}
\caption{Comparison with state-of-the-art NAS methods on NASBench-201.}
\label{tab:nb201Results}
\end{table*}

\subsubsection{Ranking Correlation} Reliable ranking performance is crucial to NAS algorithms. 
We compare the ranking performance with RSPS~\cite{rsps}, GM-NAS~\cite{gmnas} among the top 50/100/150 architectures in NASBench-201 space. RSPS does not involve splitting but trains only one supernet for 300 epochs. Following GM-NAS, we split four edges based on LID similarity, leading to $4^2=16$ sub-supernets. We finetune each sub-supernet for several epochs after partition. We run each method three times and report the mean and standard deviation of correlation results in Table \ref{tab:rank_perform}. GM-NAS and NAS-LID outperform RSPS, showing that splitting the supernet does help increase the ranking correlation among top models. We can also see that NAS-LID achieves 0.466 Kendall and 0.637 Spearman correlation, much higher than GM-NAS (0.335 Kendall and 0.484 Spearman). The lower standard deviation also indicates the stability of LID-driven characterization.

\subsubsection{Performance of Derived Architectures} For GM-NAS and our NAS-LID, we fine-tune the 16 sub-supernets for 50 epochs and then apply the evolutionary algorithm to search for superior architectures based on these sub-supernets. Note that different architectures will first inherit the weights from the corresponding sub-supernet to which they belong before the validation. We conduct the evolutionary search for 50 epochs. We produce 50 promising architectures for each search epoch via the crossover, mutation, and selection steps. Fig.   \ref{fig:EASearchNB201} compares the mean and standard deviation of validation accuracy of the searched architectures in each evolution epoch. Compared to the dramatic fluctuations in the validation accuracy of GM-NAS, we achieve higher and more stable validation performance, which further justifies that NAS-LID can better reduce the interference among architectures and thus help the training of sub-supernets. Table \ref{tab:nb201Results} compares the performance of the best architectures found by different algorithms on the CIFAR-10 and CIFAR-100 datasets. Note that the validation and test accuracy are queried from NASBench-201. The three Few-shot NAS methods adopt the same training strategy (i.e., RSPS) to train the sub-supernets and outperform RSPS by a large margin. Besides, NAS-LID can still achieve better performance and smaller variance than GM-NAS on both datasets, which confirms the advantages of LID.

\begin{table*}[!ht]
    \centering
    \begin{tabular}{l|cc|c|c|c|c}\hline
    \multirow{2}{*}{Architecture} & \multicolumn{2}{c|}{Test Error(\%)} & \multirow{2}{*}{Params (M)} & \multirow{2}{*}{Flops (M)} & \multirow{2}{*}{\makecell{Search Cost\\ (GPU Days)}} & \multirow{2}{*}{Search Method} \\\cline{2-3}
    & top-1 & top-5 & & & &  \\\hline
    \hline
    Inception-v1~\cite{inceptionv1} & 30.1 & 10.1 & 6.6 & 1448 & - & \multirow{2}{*}{manual} \\
    MobileNet~\cite{howard2017mobilenets} & 29.4 & 10.5 & 4.2 & 569 & - &  \\
    \hline
    NASNet-A~\cite{nasnet}& 26.0 & 8.4 & 5.3 & 564 & 2000 & \multirow{5}{*}{Vanilla NAS} \\
    PNAS~\cite{PNAS}& 25.8 & 8.1 & 5.1 & 588 & 225 & \\
    AmoebaNet-C~\cite{amoebanet}& 24.3 & 7.6 & 6.4 & 570 & 3150 & \\
    EfficientNet-B1~\cite{tan2019efficientnet}& 20.9 & 5.6 & 7.8 & 700 & - & \\
    MnasNet-92~\cite{tan2019mnasnet}& 25.2 & 8.0 & 4.4 & 388 & - & \\\hline
    DARTS~\cite{liu2018darts} & 26.7 & -	& 4.9 & 595 & 4 & \multirow{13}{*}{One-shot NAS}\\
    GDAS~\cite{gdas}& 26.0 & 8.5 & 5.3 & 581 & 0.3 & \\
    BayesNAS~\cite{zhou2019bayesnas}& 26.5 & 8.9 & 3.9 & - & 0.2 & \\
    P-DARTS~\cite{pdarts}& 24.4 & 7.4 & 4.9 & 557 & 0.3 & \\
    DSNAS~\cite{hu2020dsnas}& 25.7 & 8.1 & - & 324 & - & \\
    ISTA-NAS~\cite{yang2020ista-nas}& 24.0 & 7.1 & 5.7 & 638 & - & \\
    PC-DARTS~\cite{xu2019pcdarts}& 24.2 & 7.3 & 5.3 & 597 & 3.8 & \\
    BigNAS-L~\cite{yu2020bignas}& 20.5 & - & 6.4 & 586 & - & \\
    DrNAS~\cite{chen2020drnas}& 23.7 & 7.1 & 5.7 & 604 & 4.6 & \\
    SPOS~\cite{spos}& 25.3 & - & 3.4 & 328 & 8.3 & \\
    CLOSE~\cite{zhou2022closenas} & 24.7 & - & 4.8 & - & - & \\
    ProxylessNAS~\cite{cai2018proxylessnas} & 24.9 & 7.5 & 7.1 & 465 & 8.3 & \\ 
    OFA~\cite{ofanas} & 20.0 & 5.1 & 9.1 & 595 & 1.7$\ddagger$ & \\
    \hline
    
    K-Shot-NAS-A~\cite{kshotNAS} & 22.4 & 6.4 & 6.5 & 422 & 1 & \multirow{7}{*}{Few-shot NAS} \\
    FSNAS (ProxylessNAS)~\cite{fewshotNAS}& 24.1 & - & 4.9  & 521 & 20.8 & \\
    GM-NAS (ProxylessNAS)~\cite{gmnas}& 23.4 & 7.0 & 4.9 & 530 & 24.9 \\
    \textbf{NAS-LID (ProxylessNAS)} & \textbf{22.9} &  \textbf{6.3}  & 6.9 & 678 & 2.3$\dagger$ \\
    FSNAS (OFA)~\cite{fewshotNAS}&  20.2 & 5.2 & 9.2 & 600 & 1.7$\ddagger$ \\ 
    GM-NAS (OFA)~\cite{gmnas}&  19.7 & 5.0 & 9.3 & 587 & 1.7$\ddagger$ \\ 
    \textbf{NAS-LID (OFA)}& \textbf{19.5} & \textbf{5.0} & 9.9  & 776 &  1.7$\ddagger$  \\ \hline
    \end{tabular}
    \caption{Comparison with state-of-the-art NAS methods on ImageNet. $\dagger$The supernet partition and evolutionary search are conducted on the CIFAR-10 dataset. $\ddagger$ Only the search cost of evolutionary search on the ImageNet dataset is reported.}
    \label{tab:resultImagenet}
\end{table*}

\subsection{Generalizing to Other Spaces}

In addition to the tabular search space, we further evaluate NAS-LID on other open domain search spaces, including Once-for-All (OFA)~\cite{ofanas} and ProxylessNAS~\cite{cai2018proxylessnas}. We select two layers and split the supernet into four sub-supernets. We fine-tune the generated sub-supernets for 50 epochs and apply the evolutionary algorithm to search for promising architectures. We conduct the supernet partition and evolutionary search on the CIFAR-10 dataset. We then transfer the searched architectures to the ImageNet~\cite{deng2009imagenet} dataset and present the results in Table \ref{tab:resultImagenet}. We took 2.3 GPU days to split the supernet and search promising architectures on ProxylessNAS. We achieve a top-1 test error rate of 23.0\%, surpassing the one-shot and previous two few-shot methods. In the OFA space, our searched model also achieves competitive results, which validates the effectiveness of our NAS-LID. 


\subsection{LID Similarity Measures}\label{sec:sim-measure}

\begin{table}[!ht]
    \centering
    \begin{tabular}{c|cc|cc}\hline
        \multirow{2}{*}{Measure} & \multicolumn{2}{c|}{Kendall ($\tau$)} & \multicolumn{2}{c}{Spearman ($\gamma$) }\\\cline{2-5}
         & max & mean(std) & max & mean(std) \\\hline
         \hline
        Pearson & 0.30 & 0.26 (0.03) & 0.45 & 0.39 (0.04) \\\hline
        Euclidean & \textbf{0.37} &\textbf{ 0.34 (0.02)}  & \textbf{0.53} & \textbf{0.48 (0.03)} \\\hline
    \end{tabular}
    \caption{Ranking correlation of top 150 architectures on NASBench-201 with different similarity measures.}
    \label{tab:sim_measure}
\end{table}

In the above experiments, we adopt Euclidean distance to measure the LID similarity. If we ignore the scaling effect and only consider the correlations among LIDs (e.g., treating $[2, 6, 4]$ and $[20, 60, 40]$ as the same), is it also effective? To answer this, we use the Pearson correlation coefficient to measure the similarity of LIDs. As shown in Table \ref{tab:sim_measure}, Pearson similarity is not as effective as the Euclidean measure, which implies that the scale of the LID values is essential to characterize the geometric properties of different architectures in a high-dimensional space.




\section{Conclusion \& Future Work}


In this work, we empirically show that LID is a highly promising criterion for characterizing architectures, and those with high accuracy tend to have similar arch-shaped LID profiles. As a proof-of-concept, we propose NAS-LID to demonstrate a potential application of LID in splitting the supernet. Compared to the gradient, LID effectively addresses the curse of dimensionality, and the required computational overhead is naturally more negligible. NAS-LID achieves better space separability and higher ranking correlation of subnets than the gradient-based method. 

We find that the scale of LID values plays a vital role in the LID-based architecture characterization. We provide an initial intuitive understanding that the LID profile portrays how each model layer transforms the high-dimensional data into low-dimensional sub-manifolds.  Here, we only explore image classification models, but one could also study LID profiles of other models, e.g., large language model, which encourages the community to have a deeper understanding of the relationship between the LID profiles and the model generalization ability, and its broad potentials.

\paragraph{Acknowledgement.} We thank the NVIDIA Academic Hardware Grant Program for its support of our work.

\bibliography{aaai23}

\begin{thebibliography}{55}
\providecommand{\natexlab}[1]{#1}

\bibitem[{Aghajanyan, Gupta, and Zettlemoyer(2021)}]{id-explain-lmf}
Aghajanyan, A.; Gupta, S.; and Zettlemoyer, L. 2021.
\newblock Intrinsic Dimensionality Explains the Effectiveness of Language Model
  Fine-Tuning.
\newblock In \emph{Proceedings of the 59th Annual Meeting of the Association
  for Computational Linguistics and the 11th International Joint Conference on
  Natural Language Processing}, 7319--7328. Online: Association for
  Computational Linguistics.

\bibitem[{Amsaleg et~al.(2015)Amsaleg, Chelly, Furon, Girard, Houle,
  Kawarabayashi, and Nett}]{estimateLID}
Amsaleg, L.; Chelly, O.; Furon, T.; Girard, S.; Houle, M.~E.; Kawarabayashi,
  K.-i.; and Nett, M. 2015.
\newblock Estimating local intrinsic dimensionality.
\newblock In \emph{Proceedings of the 21th ACM SIGKDD International Conference
  on Knowledge Discovery and Data Mining}, 29--38.

\bibitem[{Ansuini et~al.(2019)Ansuini, Laio, Macke, and Zoccolan}]{id-profile}
Ansuini, A.; Laio, A.; Macke, J.~H.; and Zoccolan, D. 2019.
\newblock Intrinsic dimension of data representations in deep neural networks.
\newblock \emph{Advances in Neural Information Processing Systems}, 32.

\bibitem[{Bashivan, Tensen, and DiCarlo(2019)}]{bashivan2019teachernas}
Bashivan, P.; Tensen, M.; and DiCarlo, J.~J. 2019.
\newblock Teacher guided architecture search.
\newblock In \emph{Proceedings of the IEEE/CVF International Conference on
  Computer Vision}, 5320--5329.

\bibitem[{Bender et~al.(2018)Bender, Kindermans, Zoph, Vasudevan, and
  Le}]{bender2018understanding}
Bender, G.; Kindermans, P.-J.; Zoph, B.; Vasudevan, V.; and Le, Q. 2018.
\newblock Understanding and simplifying one-shot architecture search.
\newblock In \emph{International conference on machine learning}, 550--559.
  PMLR.

\bibitem[{Boykov and Jolly(2001)}]{mincut}
Boykov, Y.~Y.; and Jolly, M.-P. 2001.
\newblock Interactive graph cuts for optimal boundary \& region segmentation of
  objects in ND images.
\newblock In \emph{Proceedings eighth IEEE international conference on computer
  vision. ICCV 2001}, volume~1, 105--112. IEEE.

\bibitem[{Cai et~al.(2020)Cai, Gan, Wang, Zhang, and Han}]{ofanas}
Cai, H.; Gan, C.; Wang, T.; Zhang, Z.; and Han, S. 2020.
\newblock Once-for-All: Train One Network and Specialize it for Efficient
  Deployment.
\newblock In \emph{International Conference on Learning Representations}.

\bibitem[{Cai, Zhu, and Han(2019)}]{cai2018proxylessnas}
Cai, H.; Zhu, L.; and Han, S. 2019.
\newblock Proxyless{NAS}: Direct Neural Architecture Search on Target Task and
  Hardware.
\newblock In \emph{International Conference on Learning Representations}.

\bibitem[{Campadelli et~al.(2015)Campadelli, Casiraghi, Ceruti, and
  Rozza}]{id-benchmark}
Campadelli, P.; Casiraghi, E.; Ceruti, C.; and Rozza, A. 2015.
\newblock Intrinsic dimension estimation: Relevant techniques and a benchmark
  framework.
\newblock \emph{Mathematical Problems in Engineering}, 2015.

\bibitem[{Chen et~al.(2021{\natexlab{a}})Chen, Wang, Cheng, Tang, and
  Hsieh}]{chen2020drnas}
Chen, X.; Wang, R.; Cheng, M.; Tang, X.; and Hsieh, C.-J. 2021{\natexlab{a}}.
\newblock DrNAS: Dirichlet Neural Architecture Search.
\newblock In \emph{International Conference on Learning Representations}.

\bibitem[{Chen et~al.(2021{\natexlab{b}})Chen, Xie, Wu, and Tian}]{pdarts}
Chen, X.; Xie, L.; Wu, J.; and Tian, Q. 2021{\natexlab{b}}.
\newblock Progressive darts: Bridging the optimization gap for nas in the wild.
\newblock \emph{International Journal of Computer Vision}, 129(3): 638--655.

\bibitem[{Deng et~al.(2009)Deng, Dong, Socher, Li, Li, and
  Fei-Fei}]{deng2009imagenet}
Deng, J.; Dong, W.; Socher, R.; Li, L.-J.; Li, K.; and Fei-Fei, L. 2009.
\newblock Imagenet: A large-scale hierarchical image database.
\newblock In \emph{2009 IEEE conference on computer vision and pattern
  recognition}, 248--255. Ieee.

\bibitem[{Dong and Yang(2019{\natexlab{a}})}]{setn}
Dong, X.; and Yang, Y. 2019{\natexlab{a}}.
\newblock One-shot neural architecture search via self-evaluated template
  network.
\newblock In \emph{Proceedings of the IEEE/CVF International Conference on
  Computer Vision}, 3681--3690.

\bibitem[{Dong and Yang(2019{\natexlab{b}})}]{gdas}
Dong, X.; and Yang, Y. 2019{\natexlab{b}}.
\newblock Searching for a robust neural architecture in four gpu hours.
\newblock In \emph{Proceedings of the IEEE/CVF Conference on Computer Vision
  and Pattern Recognition}, 1761--1770.

\bibitem[{Dong and Yang(2020)}]{nb201}
Dong, X.; and Yang, Y. 2020.
\newblock NAS-Bench-201: Extending the Scope of Reproducible Neural
  Architecture Search.
\newblock In \emph{International Conference on Learning Representations}.

\bibitem[{Elsken, Metzen, and Hutter(2019)}]{nas-survey}
Elsken, T.; Metzen, J.~H.; and Hutter, F. 2019.
\newblock Neural architecture search: A survey.
\newblock \emph{The Journal of Machine Learning Research}, 20(1): 1997--2017.

\bibitem[{Guo et~al.(2020)Guo, Zhang, Mu, Heng, Liu, Wei, and Sun}]{spos}
Guo, Z.; Zhang, X.; Mu, H.; Heng, W.; Liu, Z.; Wei, Y.; and Sun, J. 2020.
\newblock Single path one-shot neural architecture search with uniform
  sampling.
\newblock In \emph{European conference on computer vision}, 544--560. Springer.

\bibitem[{He et~al.(2016)He, Zhang, Ren, and Sun}]{resnet}
He, K.; Zhang, X.; Ren, S.; and Sun, J. 2016.
\newblock Deep residual learning for image recognition.
\newblock In \emph{Proceedings of the IEEE conference on computer vision and
  pattern recognition}, 770--778.

\bibitem[{He et~al.(2021)He, Wang, Chu, Shi, Tang, Liu, Yan, Zhang, and
  Ding}]{he2021covidnet}
He, X.; Wang, S.; Chu, X.; Shi, S.; Tang, J.; Liu, X.; Yan, C.~C.; Zhang, J.;
  and Ding, G. 2021.
\newblock Automated Model Design and Benchmarking of 3D Deep Learning Models
  for COVID-19 Detection with Chest CT Scans.
\newblock In \emph{AAAI}.

\bibitem[{He et~al.(2022)He, Ying, Zhang, and Chu}]{he2022emars}
He, X.; Ying, G.; Zhang, J.; and Chu, X. 2022.
\newblock Evolutionary Multi-objective Architecture Search Framework:
  Application to COVID-19 3D CT Classification.
\newblock In \emph{International Conference on Medical Image Computing and
  Computer-Assisted Intervention}, 560--570. Springer.

\bibitem[{He, Zhao, and Chu(2021)}]{he2021automl}
He, X.; Zhao, K.; and Chu, X. 2021.
\newblock AutoML: A survey of the state-of-the-art.
\newblock \emph{Knowledge-Based Systems}, 212: 106622.

\bibitem[{Hinton et~al.(2015)Hinton, Vinyals, Dean et~al.}]{kd_hinton}
Hinton, G.; Vinyals, O.; Dean, J.; et~al. 2015.
\newblock Distilling the knowledge in a neural network.
\newblock \emph{arXiv preprint arXiv:1503.02531}, 2(7).

\bibitem[{Houle(2017)}]{houleLID1}
Houle, M.~E. 2017.
\newblock Local intrinsic dimensionality I: an extreme-value-theoretic
  foundation for similarity applications.
\newblock In \emph{International Conference on Similarity Search and
  Applications}, 64--79. Springer.

\bibitem[{Howard et~al.(2017)Howard, Zhu, Chen, Kalenichenko, Wang, Weyand,
  Andreetto, and Adam}]{howard2017mobilenets}
Howard, A.~G.; Zhu, M.; Chen, B.; Kalenichenko, D.; Wang, W.; Weyand, T.;
  Andreetto, M.; and Adam, H. 2017.
\newblock Mobilenets: Efficient convolutional neural networks for mobile vision
  applications.
\newblock \emph{arXiv preprint arXiv:1704.04861}.

\bibitem[{Hu et~al.(2022)Hu, Wang, HONG, Li, Hsieh, and Feng}]{gmnas}
Hu, S.; Wang, R.; HONG, L.; Li, Z.; Hsieh, C.-J.; and Feng, J. 2022.
\newblock Generalizing Few-Shot {NAS} with Gradient Matching.
\newblock In \emph{International Conference on Learning Representations}.

\bibitem[{Hu et~al.(2020)Hu, Xie, Zheng, Liu, Shi, Liu, and Lin}]{hu2020dsnas}
Hu, S.; Xie, S.; Zheng, H.; Liu, C.; Shi, J.; Liu, X.; and Lin, D. 2020.
\newblock Dsnas: Direct neural architecture search without parameter
  retraining.
\newblock In \emph{Proceedings of the IEEE/CVF Conference on Computer Vision
  and Pattern Recognition}, 12084--12092.

\bibitem[{Krizhevsky, Hinton et~al.(2009)}]{cifar10}
Krizhevsky, A.; Hinton, G.; et~al. 2009.
\newblock Learning multiple layers of features from tiny images.

\bibitem[{Lei et~al.(2020)Lei, An, Guo, Su, Liu, Luo, Yau, and
  Gu}]{lei2020geometric}
Lei, N.; An, D.; Guo, Y.; Su, K.; Liu, S.; Luo, Z.; Yau, S.-T.; and Gu, X.
  2020.
\newblock A geometric understanding of deep learning.
\newblock \emph{Engineering}, 6(3): 361--374.

\bibitem[{Li et~al.(2018)Li, Farkhoor, Liu, and Yosinski}]{li_id_2018_ICLR}
Li, C.; Farkhoor, H.; Liu, R.; and Yosinski, J. 2018.
\newblock Measuring the Intrinsic Dimension of Objective Landscapes.
\newblock In \emph{International Conference on Learning Representations}.

\bibitem[{Li and Talwalkar(2020)}]{rsps}
Li, L.; and Talwalkar, A. 2020.
\newblock Random search and reproducibility for neural architecture search.
\newblock In \emph{Uncertainty in artificial intelligence}, 367--377. PMLR.

\bibitem[{Liu et~al.(2019)Liu, Chen, Schroff, Adam, Hua, Yuille, and
  Fei-Fei}]{liu2019autodeeplab}
Liu, C.; Chen, L.-C.; Schroff, F.; Adam, H.; Hua, W.; Yuille, A.~L.; and
  Fei-Fei, L. 2019.
\newblock Auto-deeplab: Hierarchical neural architecture search for semantic
  image segmentation.
\newblock In \emph{Proceedings of the IEEE/CVF conference on computer vision
  and pattern recognition}, 82--92.

\bibitem[{Liu et~al.(2018)Liu, Zoph, Neumann, Shlens, Hua, Li, Fei-Fei, Yuille,
  Huang, and Murphy}]{PNAS}
Liu, C.; Zoph, B.; Neumann, M.; Shlens, J.; Hua, W.; Li, L.-J.; Fei-Fei, L.;
  Yuille, A.; Huang, J.; and Murphy, K. 2018.
\newblock Progressive neural architecture search.
\newblock In \emph{Proceedings of the European conference on computer vision
  (ECCV)}, 19--34.

\bibitem[{Liu, Simonyan, and Yang(2019)}]{liu2018darts}
Liu, H.; Simonyan, K.; and Yang, Y. 2019.
\newblock {DARTS}: Differentiable Architecture Search.
\newblock In \emph{International Conference on Learning Representations}.

\bibitem[{Liu et~al.(2020)Liu, Jia, Tan, Vemulapalli, Zhu, Green, and
  Wang}]{searchtodistill}
Liu, Y.; Jia, X.; Tan, M.; Vemulapalli, R.; Zhu, Y.; Green, B.; and Wang, X.
  2020.
\newblock Search to distill: Pearls are everywhere but not the eyes.
\newblock In \emph{Proceedings of the IEEE/CVF conference on computer vision
  and pattern recognition}, 7539--7548.

\bibitem[{Peng et~al.(2021)Peng, Zhang, Li, Wang, Liang, and Lin}]{pinas}
Peng, J.; Zhang, J.; Li, C.; Wang, G.; Liang, X.; and Lin, L. 2021.
\newblock Pi-NAS: Improving neural architecture search by reducing supernet
  training consistency shift.
\newblock In \emph{Proceedings of the IEEE/CVF International Conference on
  Computer Vision}, 12354--12364.

\bibitem[{Pham et~al.(2018)Pham, Guan, Zoph, Le, and Dean}]{enas}
Pham, H.; Guan, M.; Zoph, B.; Le, Q.; and Dean, J. 2018.
\newblock Efficient neural architecture search via parameters sharing.
\newblock In \emph{International conference on machine learning}, 4095--4104.
  PMLR.

\bibitem[{Real et~al.(2019)Real, Aggarwal, Huang, and Le}]{amoebanet}
Real, E.; Aggarwal, A.; Huang, Y.; and Le, Q.~V. 2019.
\newblock Regularized evolution for image classifier architecture search.
\newblock In \emph{Proceedings of the aaai conference on artificial
  intelligence}, volume~33, 4780--4789.

\bibitem[{Shwartz-Ziv and Tishby(2017)}]{shwartz2017opening}
Shwartz-Ziv, R.; and Tishby, N. 2017.
\newblock Opening the black box of deep neural networks via information.
\newblock \emph{arXiv preprint arXiv:1703.00810}.

\bibitem[{Su et~al.(2021)Su, You, Zheng, Wang, Qian, Zhang, and Xu}]{kshotNAS}
Su, X.; You, S.; Zheng, M.; Wang, F.; Qian, C.; Zhang, C.; and Xu, C. 2021.
\newblock K-shot NAS: Learnable Weight-Sharing for NAS with K-shot Supernets.
\newblock In \emph{International Conference on Machine Learning}, 9880--9890.
  PMLR.

\bibitem[{Szegedy et~al.(2015)Szegedy, Liu, Jia, Sermanet, Reed, Anguelov,
  Erhan, Vanhoucke, and Rabinovich}]{inceptionv1}
Szegedy, C.; Liu, W.; Jia, Y.; Sermanet, P.; Reed, S.; Anguelov, D.; Erhan, D.;
  Vanhoucke, V.; and Rabinovich, A. 2015.
\newblock Going deeper with convolutions.
\newblock In \emph{Proceedings of the IEEE conference on computer vision and
  pattern recognition}, 1--9.

\bibitem[{Tan et~al.(2019)Tan, Chen, Pang, Vasudevan, Sandler, Howard, and
  Le}]{tan2019mnasnet}
Tan, M.; Chen, B.; Pang, R.; Vasudevan, V.; Sandler, M.; Howard, A.; and Le,
  Q.~V. 2019.
\newblock Mnasnet: Platform-aware neural architecture search for mobile.
\newblock In \emph{Proceedings of the IEEE/CVF Conference on Computer Vision
  and Pattern Recognition}, 2820--2828.

\bibitem[{Tan and Le(2019)}]{tan2019efficientnet}
Tan, M.; and Le, Q. 2019.
\newblock Efficientnet: Rethinking model scaling for convolutional neural
  networks.
\newblock In \emph{International conference on machine learning}, 6105--6114.
  PMLR.

\bibitem[{Tishby, Pereira, and Bialek(2000)}]{tishby2000information}
Tishby, N.; Pereira, F.~C.; and Bialek, W. 2000.
\newblock The information bottleneck method.
\newblock \emph{arXiv preprint physics/0004057}.

\bibitem[{Xu et~al.(2022{\natexlab{a}})Xu, Chen, Quek, and
  Chong}]{xu2022fedcorr}
Xu, J.; Chen, Z.; Quek, T.~Q.; and Chong, K. F.~E. 2022{\natexlab{a}}.
\newblock FedCorr: Multi-Stage Federated Learning for Label Noise Correction.
\newblock In \emph{Proceedings of the IEEE/CVF Conference on Computer Vision
  and Pattern Recognition}, 10184--10193.

\bibitem[{Xu et~al.(2022{\natexlab{b}})Xu, Tan, Song, Luo, Leng, Qin, Liu, and
  Li}]{MAGIC}
Xu, J.; Tan, X.; Song, K.; Luo, R.; Leng, Y.; Qin, T.; Liu, T.-Y.; and Li, J.
  2022{\natexlab{b}}.
\newblock Analyzing and mitigating interference in neural architecture search.
\newblock In \emph{International Conference on Machine Learning}, 24646--24662.
  PMLR.

\bibitem[{Xu et~al.(2020)Xu, Xie, Zhang, Chen, Qi, Tian, and
  Xiong}]{xu2019pcdarts}
Xu, Y.; Xie, L.; Zhang, X.; Chen, X.; Qi, G.-J.; Tian, Q.; and Xiong, H. 2020.
\newblock PC-DARTS: Partial Channel Connections for Memory-Efficient
  Architecture Search.
\newblock In \emph{International Conference on Learning Representations}.

\bibitem[{Yang et~al.(2020)Yang, Li, You, Wang, Qian, and
  Lin}]{yang2020ista-nas}
Yang, Y.; Li, H.; You, S.; Wang, F.; Qian, C.; and Lin, Z. 2020.
\newblock Ista-nas: Efficient and consistent neural architecture search by
  sparse coding.
\newblock \emph{Advances in Neural Information Processing Systems}, 33:
  10503--10513.

\bibitem[{Yao et~al.(2022)Yao, Zhang, Yao, Wang, Ma, Zhang, Chu, Ji, Jia, Shen
  et~al.}]{yao2022edgesurvey}
Yao, J.; Zhang, S.; Yao, Y.; Wang, F.; Ma, J.; Zhang, J.; Chu, Y.; Ji, L.; Jia,
  K.; Shen, T.; et~al. 2022.
\newblock Edge-Cloud Polarization and Collaboration: A Comprehensive Survey for
  AI.
\newblock \emph{IEEE Transactions on Knowledge and Data Engineering}.

\bibitem[{Ying et~al.(2022)Ying, He, Gao, Han, and Chu}]{ying2022eagan}
Ying, G.; He, X.; Gao, B.; Han, B.; and Chu, X. 2022.
\newblock EAGAN: Efficient Two-Stage Evolutionary Architecture Search for GANs.
\newblock In \emph{European Conference on Computer Vision}, 37--53. Springer.

\bibitem[{Yu et~al.(2020)Yu, Jin, Liu, Bender, Kindermans, Tan, Huang, Song,
  Pang, and Le}]{yu2020bignas}
Yu, J.; Jin, P.; Liu, H.; Bender, G.; Kindermans, P.-J.; Tan, M.; Huang, T.;
  Song, X.; Pang, R.; and Le, Q. 2020.
\newblock Bignas: Scaling up neural architecture search with big single-stage
  models.
\newblock In \emph{European Conference on Computer Vision}, 702--717. Springer.

\bibitem[{Zhao et~al.(2021)Zhao, Wang, Tian, Fonseca, and Guo}]{fewshotNAS}
Zhao, Y.; Wang, L.; Tian, Y.; Fonseca, R.; and Guo, T. 2021.
\newblock Few-shot neural architecture search.
\newblock In \emph{International Conference on Machine Learning}, 12707--12718.
  PMLR.

\bibitem[{Zhou et~al.(2019)Zhou, Yang, Wang, and Pan}]{zhou2019bayesnas}
Zhou, H.; Yang, M.; Wang, J.; and Pan, W. 2019.
\newblock Bayesnas: A bayesian approach for neural architecture search.
\newblock In \emph{International conference on machine learning}, 7603--7613.
  PMLR.

\bibitem[{Zhou et~al.(2022)Zhou, Ning, Cai, Han, Deng, Dong, Yang, and
  Wang}]{zhou2022closenas}
Zhou, Z.; Ning, X.; Cai, Y.; Han, J.; Deng, Y.; Dong, Y.; Yang, H.; and Wang,
  Y. 2022.
\newblock Close: Curriculum learning on the sharing extent towards better
  one-shot nas.
\newblock In \emph{European Conference on Computer Vision}, 578--594. Springer.

\bibitem[{Zoph and Le(2017)}]{nas2016}
Zoph, B.; and Le, Q. 2017.
\newblock Neural Architecture Search with Reinforcement Learning.
\newblock In \emph{International Conference on Learning Representations}.

\bibitem[{Zoph et~al.(2018)Zoph, Vasudevan, Shlens, and Le}]{nasnet}
Zoph, B.; Vasudevan, V.; Shlens, J.; and Le, Q.~V. 2018.
\newblock Learning transferable architectures for scalable image recognition.
\newblock In \emph{Proceedings of the IEEE conference on computer vision and
  pattern recognition}, 8697--8710.

\end{thebibliography}

\end{document}